\documentclass[conference]{IEEEtran}
\usepackage[T1]{fontenc}
\usepackage{graphicx,booktabs,amsmath,cite,url}
\newcommand{\detparams}{3.0\,M}
\newcommand{\detepochs}{40}
\newcommand{\dflbins}{16}
\newcommand{\dflcost}{40}
\newcommand{\eestd}{83.2\pms0.6}
\newcommand{\eerelu}{82.9\pms1.2}
\newcommand{\eereg}{83.0\pms1.0}
\newcommand{\readerparams}{18k}
\newcommand{\readermmac}{2.1}
\newcommand{\qatepochs}{10}
\newcommand{\calibbatches}{300}
\newcommand{\paramsmin}{4.7k}
\newcommand{\paramsmax}{240k}
\newcommand{\fpacc}{90.5\pms0.3}
\newcommand{\acceems}{90.5}
\newcommand{\accssms}{90.4}
\newcommand{\accfeq}{90.5}
\newcommand{\accteq}{88.8}
\newcommand{\acctwogap}{10}
\newcommand{\costaf}{0.7}
\newcommand{\costat}{2.5}
\newcommand{\costwt}{31}
\newcommand{\memratiofe}{58\%}
\newcommand{\kbfe}{12.1}
\newcommand{\kbte}{9.9}
\newcommand{\accbitsee}{23}
\newcommand{\accbitsfe}{19}
\newcommand{\accbitsff}{15}
\newcommand{\peakeight}{2{,}832}
\newcommand{\ncropsgolden}{9{,}850}
\newcommand{\ncropspublic}{5{,}760}
\newcommand{\ncropsinternal}{4{,}090}
\newcommand{\goldenfixed}{99.6}
\newcommand{\onea}{38.8 / 7.7}
\newcommand{\onei}{46.4 / 6.8}
\newcommand{\oneahi}{73.2 / 64.4}
\newcommand{\ncropstrain}{54.6k}
\newcommand{\ndetframes}{14.0k}
\newcommand{\nkpi}{404}
\newcommand{\kpifrac}{15\%}
\newcommand{\npubframes}{2{,}000}
\newcommand{\npubsigns}{1{,}893}
\newcommand{\nintframes}{1{,}391}
\newcommand{\nintsigns}{1{,}667}
\newcommand{\nclasses}{15}
\newcommand{\cropsize}{48}
\newcommand{\vparentparams}{1.77\,M}
\newcommand{\vreplayparams}{0.70\,M}
\newcommand{\vsysparams}{0.72\,M}
\newcommand{\ndetval}{4{,}472}
\newcommand{\nreaderval}{6{,}096}
\newcommand{\vbgall}{27.9k}
\newcommand{\vmclosesdrange}{9 to 33}
\newcommand{\vmscrcamrec}{1.42\pms0.36}
\newcommand{\vmgainonep}{0.28}
\newcommand{\vpsgainone}{0.48\pms0.21}
\newcommand{\vpsgainonep}{0.007}
\newcommand{\vpsgainrep}{0.29\pms0.40}
\newcommand{\vpsgainrepp}{0.18}
\newcommand{\veparentparams}{2.70\,M}
\newcommand{\vegainps}{0.06\pms0.08}
\newcommand{\vepsgainone}{1.05\pms0.23}
\newcommand{\vecamiter}{0.79\pms0.53}
\newcommand{\vecamiterp}{0.03}
\newcommand{\vpubspread}{1.2}
\newcommand{\vcalibframes}{200}
\newcommand{\vecascgapall}{up to 10\%}
\newcommand{\vfireppub}{82.7\pms1.0 / 85.1\pms1.2}
\newcommand{\vfirepcam}{93.1\pms0.9 / 95.2\pms0.4}
\newcommand{\vfirepcamrec}{95.2\%}
\newcommand{\vintdecodecost}{0.6}
\newcommand{\vrefpubA}{83.9\pms0.6}
\newcommand{\vrefpubR}{86.5\pms0.6}
\newcommand{\vrefcam}{92.1\pms1.0 / 94.2\pms1.1}
\newcommand{\vrefclose}{85.4\pms1.2 / 89.9\pms1.3}
\newcommand{\vsilucam}{94.0\pms0.4 / 95.8\pms0.2}
\newcommand{\vsilupubA}{83.9\pms0.5}
\newcommand{\vsilupubR}{85.5\pms0.9}
\newcommand{\vhspub}{82.7\pms0.7 / 85.0\pms0.6}
\newcommand{\vhscam}{93.5\pms0.4 / 95.5\pms0.3}
\newcommand{\vhscloseap}{86.6}
\newcommand{\vsilucloseap}{97.3}
\newcommand{\vidlepost}{6.67}
\newcommand{\vidlerise}{1.3}
\newcommand{\vrepfpms}{2.79}
\newcommand{\vrepfpmj}{13.1}
\newcommand{\vrepfpmjt}{31.8}
\newcommand{\vrepexms}{3.91}
\newcommand{\vrepexmj}{16.3}
\newcommand{\vrepexmjt}{42.5}
\newcommand{\vrepimms}{2.67}
\newcommand{\vrepimmj}{12.1}
\newcommand{\vrepimmjt}{30.0}
\newcommand{\vrepdlams}{2.06}
\newcommand{\vrepdlamj}{8.3}
\newcommand{\vrepdlamjt}{22.1}
\newcommand{\vrepfpcam}{93.5 / 95.6}
\newcommand{\vrepdlacam}{93.6 / 95.4}
\newcommand{\vparfpms}{3.01}
\newcommand{\vparfpmj}{15.5}
\newcommand{\vparfpmjt}{35.7}
\newcommand{\vparimms}{2.65}
\newcommand{\vparimmj}{13.2}
\newcommand{\vparimmjt}{31.0}
\newcommand{\vpardlams}{2.17}
\newcommand{\vpardlamj}{8.9}
\newcommand{\vpardlamjt}{23.5}
\newcommand{\vreffpms}{3.96}
\newcommand{\vreffpmj}{20.5}
\newcommand{\vreffpmjt}{47.1}
\newcommand{\vrefexms}{4.30}
\newcommand{\vrefexmj}{17.9}
\newcommand{\vrefexmjt}{46.7}
\newcommand{\vrefimms}{3.01}
\newcommand{\vrefimmj}{12.6}
\newcommand{\vrefimmjt}{32.8}
\newcommand{\vrefdlams}{3.62}
\newcommand{\vrefdlamj}{12.8}
\newcommand{\vrefdlamjt}{36.9}
\newcommand{\vreplatcut}{7\%}
\newcommand{\vrepdlacut}{5\%}
\newcommand{\vrefexreformat}{60}
\newcommand{\vrefimreformat}{27}
\newcommand{\vrepdlatimesave}{32\%}
\newcommand{\vrepdlaenergysave}{34\%}
\newcommand{\vdlascale}{0.47}
\newcommand{\vdlalossref}{10.3}
\newcommand{\vdlalossrefmap}{5.6}
\newcommand{\vdevacc}{0.6}
\newcommand{\vpiperepdlams}{9.73}
\newcommand{\vpiperepdlafps}{94.1}
\newcommand{\vpiperepdlamj}{32.9}
\newcommand{\vpiperepdlamjt}{103.8}
\newcommand{\vpiperefimfps}{101.7}
\newcommand{\vpiperefimmj}{34.9}
\newcommand{\vpipefpsround}{94}
\newcommand{\vpipepre}{2.9}
\newcommand{\vpipedecy}{1.6}
\newcommand{\vcpurepeight}{29.1}
\newcommand{\vcamfps}{30}
\newcommand{\vcamrepms}{10.3}
\newcommand{\vreduction}{60.3\%}
\newcommand{\vsteps}{six}
\newcommand{\vstepepochs}{24}
\newcommand{\vfinalepochs}{30}
\newcommand{\vreplayepochs}{174}
\newcommand{\vperstepepochs}{120}
\newcommand{\vemparentval}{71.2}
\newcommand{\vemrepval}{71.9}
\newcommand{\vkdalpha}{0.2}
\newcommand{\vparentepochs}{100}
\newcommand{\vconvpubcost}{1.2}
\newcommand{\vconvcamcost}{0.5}
\newcommand{\pms}{$\pm$}

\begin{document}
\title{Replay Pruning and Few-Bit Integer Inference\\for Hardware-Efficient Speed-Sign Recognition}
\author{\IEEEauthorblockN{Se Un Park\IEEEauthorrefmark{1}, Kyungkeon Chung\IEEEauthorrefmark{1},  Huigeol Kim\IEEEauthorrefmark{2}, and Junyoung Park\IEEEauthorrefmark{1}\IEEEauthorrefmark{3}}\IEEEauthorblockA{\IEEEauthorrefmark{1}UX Factory, Inc. \qquad\IEEEauthorrefmark{2}BTree Co., Ltd.\qquad\IEEEauthorrefmark{3}Corresponding author}}
\maketitle

\noindent{\footnotesize\itshape This work has been submitted to the IEEE for possible publication. Copyright may be transferred without notice, after which this version may no longer be accessible.\par}\medskip
\begin{abstract}
Speed-limit sign recognition on an automotive camera system-on-chip (SoC) must fit a detector and a classifier (the reader)
into the memory, arithmetic, and energy budgets of the device. In our two-stage recognizer, the detector, a YOLO-family
network with ReLU and hard-sigmoid operators, is compressed by channel replay: an iterative pruning cascade identifies the channels to remove, the recorded
deletions are replayed on the original task-trained network, the parent, and one continuous knowledge-distillation
recovery follows. On a YOLOv5n, replay removes \vreduction{} of the parameters at the accuracy of the parent and, in validation
mAP50--95 over five matched seeds, outperforms iterative pruning and training from scratch, and its per-step variant (per-step
replay) is the most accurate pruning method. On a YOLOv8n, replay and per-step replay outperform every other method and the unpruned parent. The reader, a five-layer
CNN with \readerparams{} parameters, and the post-processing of the YOLOv5n and YOLOv8n heads use integer arithmetic
only. With 4-bit weights and 8-bit activations (W4A8), the reader matches FP32 accuracy in \kbfe\,KB of on-chip
memory, and an executable streaming model reproduces the INT8 class predictions of its runtime on all \ncropsgolden{} test crops. On
a Jetson AGX Orin, INT8 is faster than FP16 only in the implicit formulation of TensorRT, and the replay detector
runs fastest on the deep-learning accelerator (DLA). Trained on public data only, the combined system of
\vsysparams{} parameters recognizes \vfirepcamrec{} of the signs of a held-out automotive camera at 90\% precision
and \vpipefpsround{} frames per second.
\end{abstract}


\section{Introduction}
\label{sec:intro}
Speed-limit sign recognition runs on the camera system-on-chip (SoC) of a vehicle, next to the image signal processor
(ISP). A single-stage YOLO detector that localizes and classifies the signs in one pass needs megabytes of weights and
billions of operations per frame, although the task decomposes into detection over the full frame and classification of a crop a few dozen pixels wide into \nclasses{} classes. A two-stage recognizer exploits this decomposition, and its
two stages call for different reductions.

The detector carries almost all of the computation and remains on the accelerator of the SoC. We replace the SiLU
activations and the sigmoid of YOLOv5n~\cite{jocher2020yolov5} by ReLU and the hard sigmoid~\cite{howard2019searching}
and compress the network structurally. Structured pruning can serve as a search for a compact channel
configuration~\cite{liu2019rethinking}, and the weights from which a pruned network is retrained affect its
accuracy~\cite{frankle2019lottery,renda2020comparing}. We therefore use iterative pruning only to identify the
channels to remove, replay the recorded deletions on the task-trained parent, and recover the network in one
continuous run with knowledge distillation (channel replay). The reader is small enough to
leave the accelerator. We design it for placement beside the ISP, where it would process the rows of a crop as they arrive, with integer arithmetic and no
DRAM traffic, and ask how few bits it needs before a circuit implementation. Since operation
counts predict neither latency nor energy, we measure YOLOv5n and YOLOv8n on a Jetson AGX Orin across FP16, two
TensorRT formulations of INT8, and the deep-learning accelerator (DLA). Every accuracy follows a cross-camera protocol, with the networks trained on public data and our own automotive
camera held out from training.

\textbf{Related work.} Traffic-sign classifiers run on FPGAs and
microcontrollers~\cite{wong2018micronnet,kim2022lowcost,kok2025dynamic}, and a class-agnostic detector with a crop
classifier benefits fine-grained sign classes~\cite{ertler2020mapillary}. Channel pruning with Taylor
importance~\cite{li2017pruning,molchanov2019importance}, dependency graphs~\cite{fang2023depgraph}, and isomorphic
groups~\cite{fang2024isomorphic} reduces the dense computation that accelerators execute, also in YOLO detectors~\cite{zhu2025lightweight}. Integer-only inference~\cite{jacob2018quantization,yao2021hawqv3} leaves the
softmax of box decoding~\cite{li2020generalized} and the division of the IoU test in the post-processing. Closest to
our reader are low-bit quantization-aware training~\cite{esser2020learned,choi2018pact}, streaming line-buffer
dataflows on FPGAs~\cite{umuroglu2017finn,kang2023aocstream}, and bit-accurate reference models for hardware
flows~\cite{conti2020nemo,shanker2026integer}. The deployment follows Jetson Orin
benchmarks~\cite{aljami2026benchmarking} and the GPU--DLA partitioning of TensorRT~\cite{nvidia2023tensorrt,zhou2024multi}.

\textbf{Contributions.} (1) Channel replay, a pruning procedure that removes \vreduction{} of the parameters of a converted YOLOv5n and
keeps the accuracy of the original network, in matched five-seed comparisons (Section~\ref{sec:detector}). (2) A precision study and an executable integer reference model of a \kbfe-KB reader, verified against its runtime, and integer-only, division-free post-processing for the YOLOv5n and YOLOv8n heads
(Section~\ref{sec:reader}). (3) A deployment study on the Jetson AGX Orin and end-to-end results
(Sections~\ref{sec:deploy} and~\ref{sec:results}).

\section{System and Data}
\label{sec:system}
\textbf{Two-stage recognizer.} A detector localizes every speed-limit sign as a single class, a square window of
$1.3\times$ the longer side of each predicted box is scaled to $\cropsize{\times}\cropsize$ by a fixed-point bilinear
scaler, and a classifier, the reader, determines the speed value from the crop.
The combined system is the replay-compressed YOLOv5n followed by the W4A8 reader (4-bit weights, 8-bit activations).
After pruning, the detector is quantized to INT8, the lowest precision of the GPU and the DLA that run it, by
post-training quantization (PTQ) with one scale per weight channel and one per activation tensor, the maximum
absolute values over \vcalibframes{} training images.

\textbf{Dataset.} The protocol is cross-camera. Training and validation use public data only,
GTSRB~\cite{stallkamp2012man}, a Roboflow set~\cite{roboflow2023korea}, and the AIHub Korean driving
data~\cite{aihub2021seoul}, curated into \ncropstrain{} reader crops (\vbgall{} of them
background crops without a speed-limit sign) and \ndetframes{} detector frames.\footnote{The detector uses AIHub frames only (\ndetval{} validation frames). The reader is trained on GTSRB,
Roboflow, and AIHub crops (AIHub crops of other signs as background, the negative class), validated on \nreaderval{} AIHub crops, and
tested on \ncropspublic{} AIHub crops and \ncropsinternal{} crops of the held-out camera.} The public test set is the AIHub validation split (\npubframes{} frames,
\npubsigns{} signs). A close-up set of \nkpi{} images cut from those frames (sign at least \kpifrac{} of the image)
tests large, near signs. The frames of our own automotive camera (\nintframes{} test frames,
\nintsigns{} signs) are used for testing only, and, for every system, a sign counts as recognized when its number is read.\footnote{The public sets hold
no Korean minimum-speed signs, so the reader's minimum-limit classes are untrained and a minimum-speed sign is read by its number.}

\textbf{Metrics.} Detectors alone are compared by mAP50--95 on the validation split. End to end, a prediction is
correct if its box matches a labeled sign at IoU $\ge 0.5$ and its speed value is correct, and we report mAP50 and
Recognized, the recall at precision $\ge 0.9$ on the pooled precision--recall curve.

\section{Detector: Operator Conversion and Channel Replay}
\label{sec:detector}
\begin{table}[t]
\caption{Compression of the converted YOLOv5n on our dataset, with validation mAP50--95, end-to-end mAP50 / Recognized (\%), and paired differences.}
\label{tab:replay}
\centering\scriptsize
\setlength{\tabcolsep}{2pt}
\resizebox{\columnwidth}{!}{%
\begin{tabular}{@{}lccccc@{}}
\toprule
 & & & & \multicolumn{2}{c}{Minus the row, val.\ mAP50--95} \\
\cmidrule(l){5-6}
Method & \shortstack{Val.\\mAP50--95} & Public & \shortstack{Held-out\\camera} & Replay & Per-step replay \\
\midrule
Parent & 70.7 & 82.7 / 85.0 & \textbf{93.5} / \textbf{95.5} & +0.07\pms1.15 (4/5) & +0.36\pms0.87 (4/5) \\
Iterative & 70.1 & 82.7 / 84.8 & 93.3 / 95.3 & +0.63\pms0.27 (5/5)$^{\ast}$ & +0.92\pms0.64 (5/5)$^{\ast}$ \\
One-shot & 70.5 & \textbf{82.9} / 85.4 & 93.1 / 95.4 & +0.19\pms0.33 (4/5) & +0.48\pms0.21 (5/5)$^{\ast}$ \\
Replay & 70.7 & 82.6 / 85.6 & 93.1 / 95.3 & -- & +0.29\pms0.40 (4/5) \\
Per-step replay & \textbf{71.0} & 82.8 / \textbf{85.8} & 93.2 / \textbf{95.5} & $-$0.29\pms0.40 (1/5) & -- \\
Scratch & 70.2 & 82.4 / 84.8 & 92.3 / 93.9 & +0.51\pms0.28 (5/5)$^{\ast}$ & +0.80\pms0.61 (5/5)$^{\ast}$ \\
\bottomrule
\end{tabular}}
\par\vspace{2pt}{\raggedright\scriptsize Five parents pruned from \vparentparams{} to \vreplayparams{} parameters (\vreduction{} fewer). Means over five seeds with the W4A8 reader, every detector evaluated as trained in FP32 with float decoding (the deployed system quantizes the replay detector to INT8, Table~\ref{tab:systems}). Bold marks the best of the column. The last two columns are replay and per-step replay minus the method of the row in validation mAP50--95, mean $\pm$ standard deviation (points) and the seeds on which the first is higher. $^{\ast}$ marks $p<0.05$ (paired t-test).\par}
\end{table}

\textbf{Operator conversion.} Apart from their activations and heads, YOLOv5n and YOLOv8n~\cite{ultralytics2023yolov8}
consist of convolutions, additions, concatenations, upsampling, and max pooling, whereas
YOLO11~\cite{khanam2024yolov11} and YOLO26~\cite{sapkota2025ultralytics} contain a self-attention block whose softmax
needs exponentials and a division per row and are therefore not used. The SiLU and the sigmoid need lookup tables or piecewise approximations in
low precision. ReLU and the hard sigmoid, $\min(\max(x/6+1/2,\,0),\,1)$, are piecewise linear. In the YOLOv5n every
SiLU is replaced by ReLU and the sigmoid of the head by the hard sigmoid. In the YOLOv8n every SiLU is replaced by
ReLU and the distribution-focal box head~\cite{li2020generalized} by a plain regression head
(Section~\ref{sec:reader}), after which both detectors and their post-processing need only integer multiplications,
additions, shifts, and comparisons. The converted networks are trained from the COCO checkpoints with the released
recipes, as are the unconverted networks they are compared with. The conversion costs the YOLOv5n \vconvpubcost{} points of mAP50 on the public set and
\vconvcamcost{} on the held-out camera (Table~\ref{tab:systems}), changes the end-to-end mAP50 of the YOLOv8n by less
than the seed spread, and is a precondition for the DLA, on which the SiLU network runs 2.8 times slower and loses
about 17 points of mAP50.

\textbf{Channel replay.} Channel replay uses iterative pruning only to select the channels to remove and rebuilds the
compact network from the task-trained parent. \emph{The channels are selected by iterative pruning.} Taylor-importance
channel pruning~\cite{molchanov2017pruning,molchanov2019importance} on a dependency graph~\cite{fang2023depgraph},
ranking channels only against channels of the same structure~\cite{fang2024isomorphic}, is applied to the converted
YOLOv5n in \vsteps{} steps down to about 0.7\,M parameters, each followed by \vstepepochs{} epochs of recovery with
knowledge distillation (KD)~\cite{hinton2015distilling} from the full-size parent (with weight \vkdalpha, clipped to
the task loss) and \vfinalepochs{} further epochs after the last. Every channel count stays a multiple of 8, the
detection head is left intact, and the channels removed at each step are recorded. \emph{The deletions are replayed
on the parent.} The weights of the iterative model are discarded, the recorded deletions are applied in order to the
task-trained parent, which keeps the parent's final weights in the surviving channels, rather than an earlier
checkpoint as in weight rewinding~\cite{renda2020comparing}, and one continuous KD recovery of
\vreplayepochs{} epochs follows. \emph{One-shot pruning}~\cite{li2017pruning} ranks the parent's channels once and
removes them in one step, and \emph{scratch training}~\cite{liu2019rethinking} trains the replayed graph from random
initialization. \emph{Per-step replay} differs from replay only in how the channels are selected. Each step applies
the deletions made so far to the original network, the unpruned parent, retrains it for \vstepepochs{} epochs, and
ranks the next channels on it, so that no ranking is computed on weights that an earlier step has retrained. The
final graph is recovered as in replay. All methods share the parent, teacher, data, recovery budget, and model
selection.\footnote{After the shared parent training (\vparentepochs{} epochs for the YOLOv5n, \detepochs{} for the
YOLOv8n), the cascade prunes in \vsteps{} steps with \vstepepochs{} epochs of recovery after each and \vfinalepochs{}
after the last, \vreplayepochs{} in total, and its result is the iterative model. A retraining serves to rank the
next step, so replay needs only the five retrainings before the last step, \vperstepepochs{} epochs, and per-step
replay runs five of its own. Both then recover for \vreplayepochs{} epochs.} All methods are also applied to five YOLOv8n parents (\veparentparams),
pruned to 40\% of their parameters with the Ultralytics trainer and its feature distillation.

\textbf{Results} (Table~\ref{tab:replay}). On the YOLOv5n, replay matches the parent with \vreduction{} fewer
parameters. It outperforms training from scratch and iterative pruning on all five seeds in validation mAP50--95 and scratch
training by \vmscrcamrec{} Recognized points on the held-out camera as well, whereas its difference from one-shot
pruning is not significant ($p=\vmgainonep$). Per-step replay
achieves the highest validation mAP50--95 of all methods, outperforms one-shot pruning on all five seeds (\vpsgainone,
$p=\vpsgainonep$), and does not differ significantly from replay (\vpsgainrep, $p=\vpsgainrepp$). On the YOLOv8n, replay outperforms iterative pruning, one-shot pruning, scratch training, and even the parent on all five seeds
(\vemrepval{} against \vemparentval{} for the parent),
although the replayed networks have \vecascgapall{} fewer parameters than the one-shot networks, and per-step
replay matches replay (\vegainps). End to end on the held-out camera, no
pair among replay, per-step replay, iterative pruning, and one-shot pruning is separated at $p<0.05$ in mAP50 or
Recognized on the YOLOv5n. On the YOLOv8n, replay outperforms iterative pruning by \vecamiter{} mAP50 points
($p=\vecamiterp$), and the other pairs are not separated. Two factors are thus separated: the recovery weights, since replay outperforms scratch training, which recovers
the same graph from random weights on the same schedule (iterative pruning differs in both weights and schedule), and the channel selection, since per-step replay and one-shot pruning recover
from the parent's weights with the same recovery budget and differ only in the channels they remove, with per-step replay ahead on all five seeds of both detectors (\vpsgainone{} and \vepsgainone{} validation points).

\section{Few-Bit Integer Reader and Integer-Only Pipeline}
\label{sec:reader}
\textbf{Reader and quantization scheme.} The reader, \readerparams{} parameters and \readermmac\,MMAC per $\cropsize{\times}\cropsize$ crop, consists of five $3{\times}3$ convolutions (8, 16, 16, 32, 32
channels, strides 2, 1, 2, 1, 2) with ReLU, a global average pool, and a fully connected layer to \nclasses{} classes
(maximum limits 30--120\,km/h, minimum limits 30--60\,km/h, and background), with three-row line buffers per layer and no stored feature map. Weights are
quantized symmetrically per output channel at $W$ bits, activations after ReLU are unsigned at $A$ bits, biases are
INT32, and requantization uses an INT32 multiplier and a shift, or a shift alone with power-of-two scales. From
the same FP32 reader, PTQ sets the clipping range of every tensor to the
range that minimizes the mean squared quantization error over \calibbatches{} batches of training crops (MSE
ranges), and quantization-aware training (QAT) fine-tunes for \qatepochs{} epochs with a straight-through
estimator~\cite{esser2020learned}.

\begin{figure}[t]
\centering\includegraphics[width=\columnwidth]{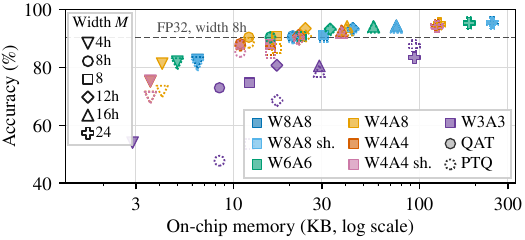}
\caption{Accuracy of the reader alone on the public test crops against its on-chip memory. Marker shape gives the width $M$
(the five layers have $M$, $2M$, $2M$, $4M$, and $8M$ channels, the suffix h halves the last layer, \paramsmin{} to
\paramsmax{} parameters) and marker color the precision, where sh.\ denotes power-of-two scales. Filled markers are QAT and hollow markers PTQ, with three seeds per point. The dashed line is the
reader of width 8h in FP32 (\fpacc\%), the deployed width.}
\label{fig:bits}
\end{figure}

\textbf{Precision study} (Fig.~\ref{fig:bits}). (i) At 8 and 6 bits PTQ is lossless (\acceems\% and \accssms\% against
\fpacc\%). (ii) Weights and activations degrade differently. With 8-bit weights, 4- and 3-bit activations cost
\costaf{} and \costat{} points, respectively, under PTQ, whereas 3-bit weights lose about \costwt{} points under PTQ
and need QAT, which recovers them to \accteq\% at \kbte\,KB. W4A8 matches FP32 (\accfeq\%) at \kbfe\,KB,
\memratiofe{} of the W8A8 memory. Two-bit weights lose \acctwogap{} points or more even with group scales and QAT.
(iii) The worst-case accumulator, bounded by the sum of $|w|$ over a kernel times the activation maximum plus the
bias, needs \accbitsee{}, \accbitsfe{}, and \accbitsff{} bits at W8A8, W4A8, and W4A4, respectively. We deploy W4A8.

\textbf{Integer reference model.} An executable Python model computes with INT32 accumulators, processes the crop row by row
with three-row line buffers, and records the bytes held at every step. Against ONNX
Runtime (8-bit models) and the trained PyTorch model (other precisions), on all \ncropsgolden{} test crops, it predicts the same class on every crop, with identical INT8 logits on all crops against PyTorch and on \goldenfixed\% against ONNX Runtime with the INT32 multiplier and shift. The line
buffers hold at most \peakeight{} bytes, and the streaming model occupies \kbfe\,KB in total at W4A8 with no DRAM traffic.

\textbf{Integer-only post-processing.}
\label{sec:intpost}
The score threshold $t$ of the YOLOv8n is applied to the logit, as $\mathrm{logit}(t)$, so that no sigmoid is
evaluated, and the IoU test of NMS is cross-multiplied. The distribution-focal head of the YOLOv8n decodes each box
side as the softmax-weighted mean of \dflbins{} bins, which needs exponentials and a division. Taking the most
probable bin instead costs \dflcost{} points of mAP50, because the box sides snap to whole cells and duplicate boxes
survive NMS. We therefore train the YOLOv8n with a regression head that predicts each side directly, at equal
accuracy (\eereg\% end-to-end mAP50 against \eerelu\% with the distribution-focal head and \eestd\% for the
unconverted YOLOv8n). End to end, the integer post-processing changes mAP50
and Recognized by at most \vintdecodecost{} points for both heads.

\section{Deployment on the Jetson AGX Orin}
\label{sec:deploy}
\begin{table}[t]
\caption{Median GPU time and energy per inference of the detector engines on the Jetson AGX Orin.}
\label{tab:jetson}
\centering\scriptsize
\setlength{\tabcolsep}{3pt}
\resizebox{\columnwidth}{!}{%
\begin{tabular}{@{}llccc@{}}
\toprule
Detector & Precision & ms & mJ, above idle & mJ, total \\
\midrule
Replay YOLOv5n (\vreplayparams) & FP16 & \vrepfpms & \vrepfpmj & \vrepfpmjt \\
 & INT8 explicit & \vrepexms & \vrepexmj & \vrepexmjt \\
 & INT8 implicit & \vrepimms & \vrepimmj & \vrepimmjt \\
 & DLA INT8 & \textbf{\vrepdlams} & \textbf{\vrepdlamj} & \textbf{\vrepdlamjt} \\
\addlinespace[2pt]
Parent YOLOv5n (\vparentparams) & FP16 & \vparfpms & \vparfpmj & \vparfpmjt \\
 & INT8 implicit & \vparimms & \vparimmj & \vparimmjt \\
 & DLA INT8 & \textbf{\vpardlams} & \textbf{\vpardlamj} & \textbf{\vpardlamjt} \\
\addlinespace[2pt]
YOLOv8n, regression head (2.70\,M) & FP16 & \vreffpms & \vreffpmj & \vreffpmjt \\
 & INT8 explicit & \vrefexms & \vrefexmj & \vrefexmjt \\
 & INT8 implicit & \textbf{\vrefimms} & \textbf{\vrefimmj} & \textbf{\vrefimmjt} \\
 & DLA INT8 & \vrefdlams & \vrefdlamj & \vrefdlamjt \\
\bottomrule
\end{tabular}}
\par\vspace{2pt}{\raggedright\scriptsize One uninterrupted run (TensorRT 8.5.2, 30\,W mode, batch 1, $640{\times}640$, seed 0). Energy
above the post-run idle (\vidlepost\,W) and in total. The DLA rows run INT8 with the head outputs in FP16. Bold marks the fastest and most efficient per detector.\par}
\end{table}

\begin{table}[t]
\caption{End-to-end mAP50 / Recognized (\%) on our dataset.}
\label{tab:systems}
\centering\scriptsize
\setlength{\tabcolsep}{3pt}
\resizebox{\columnwidth}{!}{%
\begin{tabular}{@{}lccc@{}}
\toprule
Detector & Params & Public & Held-out camera \\
\midrule
Unconverted YOLOv5n (SiLU, sigmoid) & \vparentparams & \textbf{\vsilupubA} / \vsilupubR & \textbf{\vsilucam} \\
Parent YOLOv5n (ReLU, hard sigmoid) & \vparentparams & \vhspub & \vhscam \\
Replay YOLOv5n (combined system) & \textbf{\vreplayparams} & \vfireppub & \vfirepcam \\
YOLOv8n, regression head & 2.70\,M & \textbf{\vrefpubA} / \textbf{\vrefpubR} & \vrefcam \\
Merged recognizer (YOLOv8n)$^{a}$ & \detparams & \onea & \onei \\
\bottomrule
\end{tabular}}
\par\vspace{2pt}{\raggedright\scriptsize Public training data, 640\,px, the same W4A8 reader (18k parameters, not in Params). Mean \pms{} standard deviation over five
seeds (three for the unconverted YOLOv5n and the merged recognizer). The replay YOLOv5n and YOLOv8n rows are fully integer (INT8
detector in ONNX Runtime, integer post-processing), the other rows FP32 (Table~\ref{tab:replay} gives the replay detector in FP32 with float decoding). Bold: the smallest detector and the best mean of each column (ties at this precision share it). $^{a}$One network for the sign locations
and the speed classes, no reader.\par}
\end{table}

\textbf{Protocol.} Both detector families were measured on a Jetson AGX Orin (Table~\ref{tab:jetson}) in three 30\,s windows after
a warm-up, with the median GPU time of TensorRT's command-line tool as the latency and the energy from the module's
power rails.\footnote{trtexec, whose GPU time includes the DLA execution. The energy is read from the INA3221 rails
VDD\_GPU\_SOC, VDD\_CPU\_CV, and VIN\_SYS\_5V0. All engines ran in one uninterrupted run, with the post-run idle as
the reference, since the idle power rose by \vidlerise\,W after the first engine.} The INT8 engines use
the ranges of Section~\ref{sec:system} in two TensorRT formulations, explicit with quantize--dequantize nodes in the
graph and implicit with the ranges as a calibration cache, and the DLA, a fixed-function engine beside the
GPU, runs INT8 with FP16 head outputs.

\textbf{Detectors} (Table~\ref{tab:jetson}). (i) Latency does not follow the operation count. Replay halves the convolution FLOPs of the YOLOv5n but shortens its FP16 latency by only \vreplatcut, its DLA
latency by \vrepdlacut, and its implicit-INT8 latency not at all. (ii) INT8 shortens the latency only in the implicit formulation. The explicit
engine is the slowest configuration of every detector, since its quantize--dequantize nodes fix the precision at every
tensor boundary and force reformatting layers (\vrefexreformat{} against \vrefimreformat{} in the implicit YOLOv8n
engine). The implicit engine is the fastest GPU configuration and stays within \vdevacc{} points of its accuracy on the development machine. (iii) The DLA runs each
detector as one INT8 subgraph without per-layer kernel launches, which
makes it the fastest for the two YOLOv5n detectors but slower than the implicit GPU engine for the
larger YOLOv8n. The YOLOv8n needs
its head outputs in FP16. Box distances of tens of cells and the score logits share one INT8 scale, a step of
about half a cell and of \vdlascale{} in the logits, which costs \vdlalossrefmap{} mAP50 and
\vdlalossref{} Recognized points on the public set. The bounded hard-sigmoid outputs of the YOLOv5n stay within the
seed spread in INT8. The replay YOLOv5n on the DLA is therefore the fastest and most efficient configuration and keeps its FP16 accuracy (\vrepdlacam{} against \vrepfpcam{} on the
held-out camera). Compared with the YOLOv8n reference in its best configuration, implicit INT8, it needs
\vrepdlatimesave{} less time and \vrepdlaenergysave{} less energy.

\textbf{Pipelines.} The combined system, the replay YOLOv5n on the DLA followed by the W4A8 reader as an INT8 engine
on the GPU,\footnote{Its memory differs from the \kbfe\,KB of the streaming model, which packs the weights in 4 bits
and holds three rows per layer, whereas the engine stores INT8 weights and whole feature maps.} runs at
\vpiperepdlafps{} frames per second (fps) over the whole loop with frames in memory, with a median
frame time of \vpiperepdlams\,ms, at \vpiperepdlamj\,mJ per frame above idle and \vpiperepdlamjt\,mJ in total, and the YOLOv8n reference pipeline in implicit INT8 at \vpiperefimfps{} fps and
\vpiperefimmj\,mJ above idle. The frame thus costs nearly five times the \vrepdlams\,ms and \vrepdlamjt\,mJ of the detector alone, because the input resizing (\vpipepre\,ms), the box
decoding with NMS (\vpipedecy\,ms), and the cropping run in Python on the host, each engine launch adds copies and
synchronization, and the energy includes the idle power over the frame time. A real-time program with a USB camera
viewing road footage on a screen is bound by the camera's \vcamfps{} fps and processes each frame in \vcamrepms\,ms, and without the GPU and the DLA, in ONNX Runtime on the
eight Arm cores, it runs at \vcpurepeight{} fps.

\section{End-to-End Results and Limitations}
\label{sec:results}
Table~\ref{tab:systems} compares the systems on our dataset. The fully integer combined system reaches
\vfirepcam{} with \vsysparams{} parameters.\footnote{We adopt the replay detector. Per-step replay does not differ from it
significantly (\vpsgainrep{} validation points, $p=\vpsgainrepp$).} On the public set the four
detectors lie within \vpubspread{} points of mAP50, whereas a merged recognizer, the YOLOv8n trained on the detector frames alone as sign locator
and speed classifier in one network (\detparams{} parameters), reaches only \onea{} (\oneahi{} at 1280\,px). The gain at 1280\,px indicates that the limited sign resolution at its stride-8 class head may contribute to the gap, whereas the reader classifies each sign at crop resolution.

\textbf{Limitations.} On close-up signs, which fill far more of the image than the training signs, the converted YOLOv5n detects poorly and inconsistently (detector AP50 \vhscloseap{} against \vsilucloseap{}
for the unconverted SiLU network), the compressed models inherit
this with standard deviations of \vmclosesdrange{} points across seeds, and the YOLOv8n reference reaches \vrefclose{} on them. All tests use Korean speed signs. The
device measurements use one board in its 30\,W mode, one seed for the DLA and implicit-INT8 accuracies of the
YOLOv8n, and energy from the GPU, CPU, and 5\,V rails of the module, which excludes the memory rail, the carrier board, and the supply losses at the AC input.

\section{Conclusion}
\label{sec:conclusion}
We have presented a speed-sign recognizer that pairs a YOLO-family detector with a small integer
reader. Channel replay removes \vreduction{} of the parameters of a converted YOLOv5n at the accuracy of its parent,
the \readerparams-parameter reader matches FP32 accuracy at W4A8 in a \kbfe-KB streaming model verified against its runtime, and on the Jetson AGX Orin INT8 is faster than FP16 only in the implicit formulation of TensorRT, with the replay detector fastest on the DLA. The combined system, trained on public data only, recognizes \vfirepcamrec{} of
the signs of a held-out camera at 90\% precision and \vpipefpsround{} fps. Future work will implement
the reader in RTL, train the minimum-speed classes, and reduce the detector toward microcontroller-class
hardware.

\bibliographystyle{IEEEtran}
\bibliography{refs}
\end{document}